\documentclass[sigconf,nonacm]{acmart}
\usepackage{booktabs,tabularx}
\begin{document}

\title{A Camera-Native Stereo VR180 Dataset}

\author{Linxuan Lu}
\email{linxuan.lu@qq.com}
\affiliation{%
  \institution{Independent Researcher}
  \country{Singapore}}

\renewcommand{\shortauthors}{Lu}

\begin{abstract}
Immersive VR180 video is increasingly produced with professional stereo fisheye cameras, yet public VR180 research resources are mostly collected from online platforms such as YouTube: already stitched, projected and compressed by unknown pipelines, and without lens calibration. We present a firsthand-captured stereo VR180 dataset recorded with two Blackmagic URSA Cine Immersive cameras. It contains 1,211 samples---636 stereo video clips (2,220.8\,s, mostly 90\,fps) and 575 stereo stills---each released as camera-native Blackmagic RAW, separate-eye native fisheye HEVC ($8160\times7200$ per eye) and half-equirectangular HEVC ($7200\times7200$ per eye), together with the factory lens calibration, portable fisheye/half-equirectangular conversion tools and AI-generated scene and visual-challenge annotations. Re-encoding the released fisheye and half-equirectangular renders with x265 over 24 clips, both eyes, four rate points and nine viewing directions, native-fisheye coding needed more bitrate than half-equirectangular coding at equal viewport quality for all 24 clips (median $+38\%$), in every part of the field of view. Data: \url{https://huggingface.co/datasets/lulinxuan/VR180}; code: \url{https://github.com/lulinxuan/vr180-dataset-tools}.
\end{abstract}

\ccsdesc[500]{Information systems~Multimedia information systems}
\ccsdesc[300]{Computing methodologies~Image compression}
\ccsdesc[300]{Computing methodologies~Virtual reality}

\keywords{VR180, stereoscopic video, fisheye, omnidirectional video, RAW video, dataset, video coding}

\begin{teaserfigure}
\centering
\includegraphics[width=\textwidth]{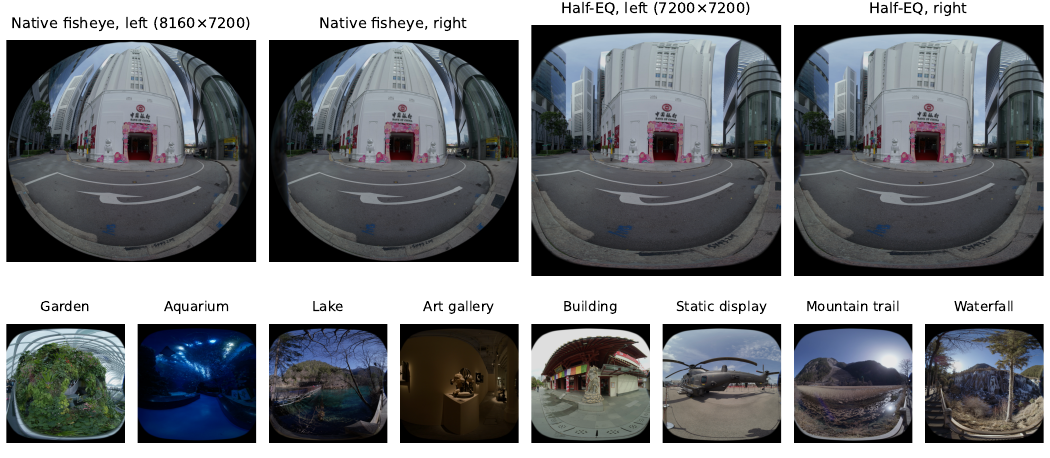}
\caption{Top: one stereo sample as released, in native fisheye and half-equirectangular (half-EQ) form for both eyes. Bottom: half-EQ left-eye frames from eight of the 41 scene types.}
\Description{Four images of the same scene: left and right native fisheye frames with a circular image area, and left and right half-equirectangular frames. Below, eight smaller frames from different scenes such as a garden, an aquarium tunnel, a city street, a lake, an art gallery, a temple building and a frozen waterfall.}
\label{fig:teaser}
\end{teaserfigure}

\maketitle

\section{Introduction}
Stereo VR180 video is captured through two side-by-side fisheye lenses about 60--64\,mm apart, close to the human interpupillary distance, each covering about 180$^\circ$, and is a common format for immersive cinema, live events and spatial-video playback. Professional VR180 cameras capture each eye through a fisheye lens, but distribution pipelines commonly convert the footage to an equirectangular layout before compression. Whether this conversion helps or hurts coding efficiency across the field of view is hard to study with existing resources: public VR180 collections are mostly gathered from online platforms such as YouTube, so their footage has already been stitched, projected and compressed by unknown pipelines, and lens calibration is unavailable.

We release a dataset captured firsthand with two cameras of one professional immersive model. Every sample is kept in its camera-native Blackmagic RAW (BRAW) form, so users can render references with known processing or their own tone mapping, and is paired with two display representations of the same rays: native fisheye and half-equirectangular (half-EQ). Factory lens data accompany each sample, and portable tools convert between the two representations within documented limits (Fig.~\ref{fig:teaser}). Our contributions are:
\begin{itemize}
\item \textbf{A camera-native stereo VR180 resource}: 1,211 samples with BRAW, per-eye fisheye ($8160\times7200$) and half-EQ ($7200\times7200$) HEVC, linked by stable identifiers.
\item \textbf{Calibration and tools}: the manufacturer's lens calibration of both cameras, and a portable fisheye$\leftrightarrow$half-EQ converter validated on Linux, with measured coverage and alignment limits.
\item \textbf{A representation coding study}: native-fisheye versus half-EQ coding of the released renders, reported per clip, per viewing direction and per content attribute; half-EQ needed less bitrate for every clip (median 38\%), although its angular sampling density is not higher.
\end{itemize}

\section{Related Work}
\textbf{VR180 and stereo video datasets.} WebVR180 curates 396,597 web-derived VR180 clip--caption pairs at $2048\times1024$ for stereoscopic video generation~\cite{webvr180}. Stereo4D reconstructs camera poses, depth and 3D motion for about 110k clips from internet VR180 videos~\cite{stereo4d}, and a quality study analyzed stereoscopic artifacts in 1,000 online VR180 videos~\cite{lavrushkin2021}. Captured stereo video and image collections include StereoV1K~\cite{spatialme}, SVD~\cite{svd} and Holopix50k~\cite{holopix50k}, which use ordinary-FoV stereo devices rather than hemispherical lenses. These resources offer scale or rich geometry, but their media are processed and compressed, and none provides camera RAW with factory lens calibration.

\textbf{Omnidirectional video datasets and coding.} D-SAV360 distributes 50 stereo and 35 mono 360$^\circ$ videos with gaze data, together with the original six-fisheye recordings~\cite{dsav360}; CIVIT distributes Nokia OZO fisheye views with stitched panoramas~\cite{civit}. MMSys has hosted head-movement~\cite{corbillon2017,lo2017} and rate--distortion~\cite{chakareski2021} datasets for 360$^\circ$ streaming, and VQA-ODV supports quality assessment~\cite{vqaodv}. Projection formats and spherical metrics for 360$^\circ$ coding are specified in JVET 360Lib~\cite{360lib}, including WS-PSNR~\cite{wspsnr}. For fisheye video, Eichenseer and Kaup found that coding distortion-corrected sequences can outperform coding the fisheye directly~\cite{eichenseer2014}. Distributing lens-domain and panoramic media together is therefore not new; what our dataset adds is a stereo VR180 source with camera RAW in which both representations are rendered from the same capture, enabling the comparison in Sec.~\ref{sec:study}. Table~\ref{tab:resources} summarizes the differences; WoodScape~\cite{woodscape} illustrates calibrated fisheye data in a different (automotive) domain.

\begin{table*}[t]
\centering\small
\caption{Scope of related resources. Units differ and counts are not a quality measure. NR: not reported in the cited documentation (not assumed absent).}
\label{tab:resources}
\begin{tabularx}{\textwidth}{@{}lll>{\raggedright\arraybackslash}Xll>{\raggedright\arraybackslash}X@{}}
\toprule
Resource & Capture & Scale & Released resolution & Camera RAW & Lens-domain views & Calibration / labels \\
\midrule
WebVR180~\cite{webvr180} & web VR180 & 396,597 clips & $2048\times1024$ & no & no & NR / captions \\
Stereo4D~\cite{stereo4d} & web VR180 & $\sim$110k clips & YouTube links; annotations on $512^2$ rectified views & no & no & estimated poses / depth, 3D tracks \\
D-SAV360~\cite{dsav360} & Kandao Obsidian & 50 stereo + 35 mono & $3840^2$ stereo ERP; $3000\times2160$ fisheye & no (H.264) & six fisheye & NR / gaze, depth estimates \\
CIVIT~\cite{civit} & Nokia OZO & 5 sequences & 2K$\times$2K per view & NR & 4 or 8 fisheye & metadata / -- \\
StereoV1K~\cite{spatialme} & Canon dual lens & 1,000 clips & $1180\times1180$ & NR & NR & NR / -- \\
SVD~\cite{svd} & iPhone, Vision Pro & 300 + 10 videos & NR & NR & no & NR / -- \\
\textbf{Ours} & URSA Cine Immersive & 636 clips + 575 stills & $8160\times7200$ fisheye, $7200^2$ half-EQ & BRAW & 2 (stereo) & factory ILPD / scene, motion, challenges \\
\bottomrule
\end{tabularx}
\end{table*}

\section{Dataset Construction}
\label{sec:construction}
\begin{figure}[t]
\centering
\includegraphics[width=\columnwidth]{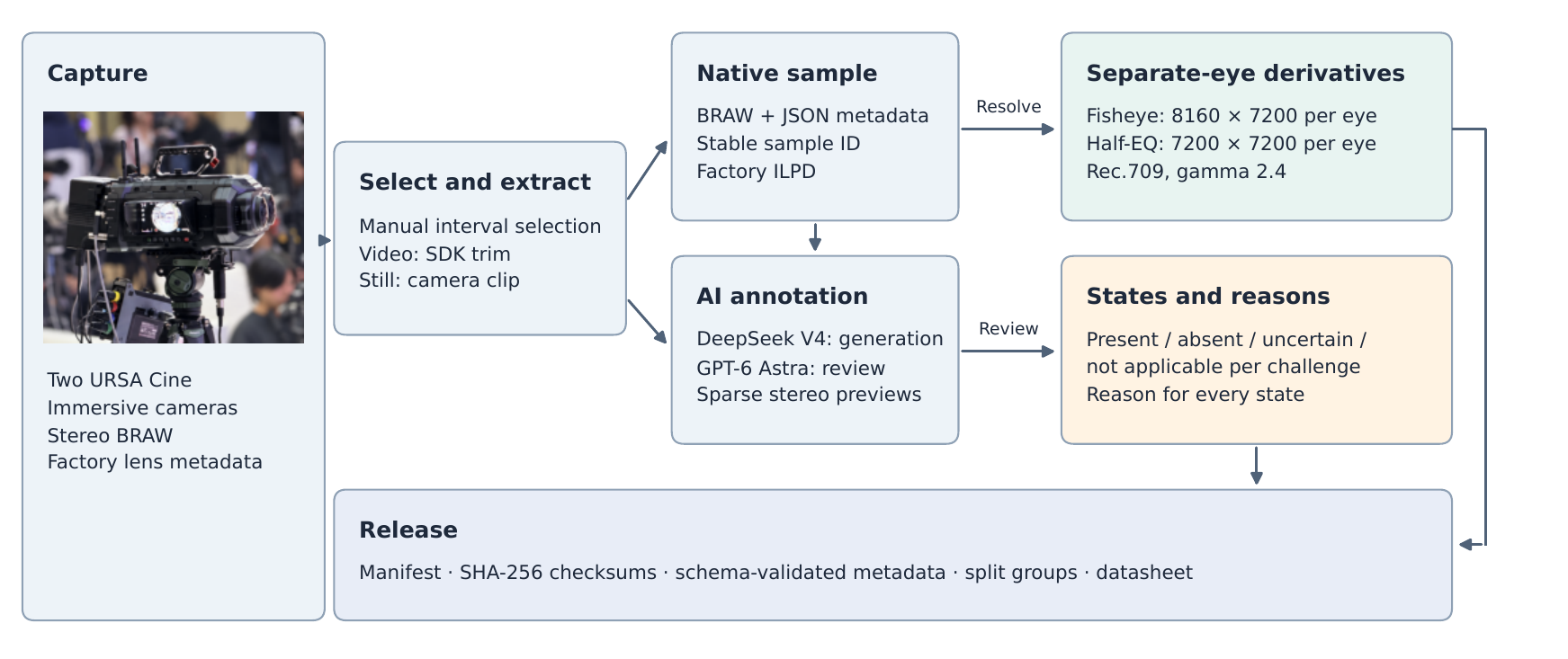}
\caption{From firsthand capture to released samples. Left: one of the two capture cameras on location.}
\Description{A flow chart. The capture box shows a photo of a Blackmagic URSA Cine Immersive camera with two fisheye lenses on a tripod at an event, its monitor displaying a fisheye image. Steps: capture with two URSA Cine Immersive cameras; selection and extraction (SDK trim for video, BRAW copy for stills); native sample with metadata; Resolve renders to fisheye and half-EQ derivatives; AI annotation followed by review and refinement; release with manifest and checks.}
\label{fig:pipeline}
\end{figure}

\textbf{Capture.} All footage was recorded by the author or by camera operators hired by the author with two Blackmagic URSA Cine Immersive cameras (Fig.~\ref{fig:pipeline}) on 30 recording dates between August 2025 and August 2026, mostly in public venues: events and exhibitions, gardens, museums, zoos and aquariums, streets, lakes and mountains (Fig.~\ref{fig:teaser}). The release is a manually selected subset of about ten hours of footage.

\textbf{Selection and native extraction.} Clips were chosen in a custom tool by selecting a contiguous interval during stereo playback; durations range from 0.4 to 11.8\,s (median 5.0\,s). Videos are cut with the Blackmagic RAW SDK trim operation, which copies compressed RAW frames without re-encoding; the tool reopens each output and checks frame count, dimensions, frame rate and lens metadata. Clips are kept short to limit data volume and were chosen to cover diverse scenes. Stereo stills are original one-frame camera clips. Selection was not randomized (Sec.~\ref{sec:limitations}).

\textbf{Representations.} Each sample comprises the BRAW file, an English JSON metadata record and two separate-eye HEVC representations (Main10, 4:2:0, QuickTime): native fisheye at $8160\times7200$ per eye and half-EQ at $7200\times7200$ per eye, where one eye covers $180^\circ\times180^\circ$. The sensor crops the lens image at the top and bottom, so about 11\% of each half-EQ frame, mostly near the zenith and nadir, is black, while the fisheye also records content beyond $90^\circ$ at the sides. Both were rendered in DaVinci Resolve to SDR Rec.709 with gamma~2.4 and carry the camera's audio where it was recorded; Sec.~\ref{sec:validation} compares their tone statistics. Because the Resolve grade is not fully reproducible, experiments that need a fully controlled reference can decode the BRAW directly.

\textbf{Lens metadata and conversion tools.} Each camera has its own factory lens calibration (an ILPD file), released unchanged with its identifier; its pixel coordinates refer to the native fisheye raster. For each camera we release per-eye lookup tables ($4096^2$, generated once with Apple's immersive-media framework) and a Python/OpenCV/FFmpeg converter between fisheye and half-EQ that runs without macOS-specific libraries.

\section{Annotation}
\label{sec:annotation}
Each sample has an English scene description, a scene type (41 values), an optional subtype (111 values; null for 388 samples), camera and object motion (videos only), and eight visual challenges (Table~\ref{tab:challenges}).

Labels were generated with the DeepSeek V4 API and reviewed with GPT-6 Astra, both from 960-px previews: six left-eye frames spread over the clip and two right-eye frames for videos, and one frame per eye for stills. Event scenes are split into six types---exhibition hall, outdoor static display, air show, vehicle show, pet show and performance---assigned by rules from the scene descriptions and checked by Claude on the preview frames. Two challenges use operational rules: \emph{low light} is defined from rendered-preview luma statistics and \emph{close-up} candidates come from sparse stereo correspondences with large disparity (definitions in the dataset's \url{docs/ANNOTATIONS.md}). No human verification was performed and we report no label accuracy. The labels are intended for browsing, filtering and stratifying the collection---as in Sec.~\ref{sec:study}---not as ground truth for training recognition models.

\begin{table}[t]
\centering\small
\caption{Challenge states over the 1,211 samples (AI-generated). Fast motion is not applicable (N/A) to stills.}
\label{tab:challenges}
\begin{tabular}{lrrrr}
\toprule
Challenge & Present & Absent & Uncertain & N/A \\
\midrule
Occlusion & 543 & 584 & 84 & 0 \\
Close-up objects & 41 & 554 & 616 & 0 \\
Reflections & 465 & 686 & 60 & 0 \\
Transparent objects & 376 & 781 & 54 & 0 \\
Low texture & 686 & 407 & 118 & 0 \\
Repetitive patterns & 572 & 559 & 80 & 0 \\
Low light & 282 & 929 & 0 & 0 \\
Fast motion & 0 & 535 & 101 & 575 \\
\bottomrule
\end{tabular}
\end{table}

\section{Dataset Statistics}
\label{sec:stats}
\begin{figure}[t]
\centering
\includegraphics[width=\columnwidth]{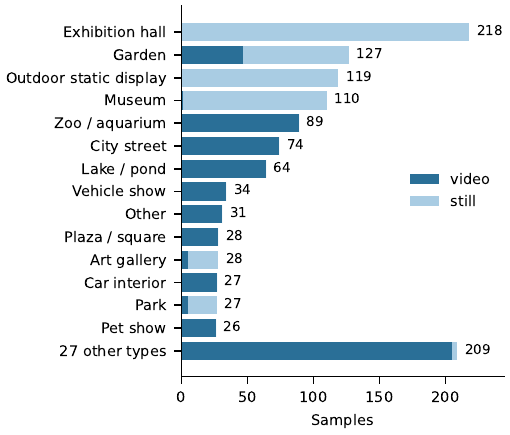}
\caption{Scene types (AI-generated labels): the 14 most frequent and the remaining 27 combined, split into videos and stills.}
\Description{A horizontal stacked bar chart of sample counts per scene type, split into videos and stills. Exhibition hall has 218 samples (all stills), followed by garden with 127, outdoor static display with 119 and museum with 110 (both all stills), zoo or aquarium with 89 and city street with 74; 27 rarer types together have 209 samples.}
\label{fig:scenes}
\end{figure}

Table~\ref{tab:stats} and Fig.~\ref{fig:scenes} summarize the collection. Scene types are imbalanced: one aviation and defense exhibition contributes 337 samples (218 exhibition hall, 119 outdoor static display), followed by gardens (127) and museums (110). Several samples can come from one recording, so random per-sample splits leak content. Camera ID, recording date and source clip number define 712 candidate recording groups (268 with more than one sample); joining groups connected by automated visual-similarity matches yields 662 overlap components, which we release in \texttt{split-groups.json}. We provide no fixed train/test split; users should assign whole components to a split.

\begin{table}[t]
\centering\small
\caption{Collection statistics (decimal units). Storage differences reflect release encoder settings, not a controlled comparison.}
\label{tab:stats}
\begin{tabular}{lr}
\toprule
Samples (videos / stills) & 1,211 (636 / 575) \\
Total / median video duration & 2,220.8\,s / 5.0\,s \\
Clips at 90 / 30 fps & 632 / 4 \\
Eye-frames per representation & 399,484 \\
Recording dates / cameras & 30 / 2 \\
BRAW storage & 5.41\,TB \\
Fisheye / half-EQ HEVC storage & 314.8 / 278.0\,GB \\
\bottomrule
\end{tabular}
\end{table}

\section{Technical Validation}
\label{sec:validation}
\textbf{Completeness and decoding.} All 1,211 samples resolve to their BRAW, metadata and four separate-eye files (4,844 HEVC files). Container probes report no missing eye and no frame-count, frame-rate or duration mismatch, and all metadata pass schema and still/video invariant checks. All 2,422 half-EQ files (399,484 eye-frames) were decoded completely in software without decoder errors, with frame counts and timestamps matching the container headers. Checksums for every file are distributed with the dataset.

\textbf{Color consistency between representations.} Because the two representations are separate renders, we compared tone statistics of frame 0 (left eye, non-black pixels) for every sample: the fisheye/half-EQ ratio of mean saturation is 0.85--1.41 (median 1.02) and that of luma standard deviation 0.75--1.17 (median 1.00).

\textbf{Left/right consistency.} On one frame per sample (the middle frame of videos), we compared the two eyes inside the central $\pm45^\circ$ of the half-EQ image (Table~\ref{tab:stereo}). The median absolute luma difference is 0.40\%, the median color difference of the mean colors $\Delta E$ is 0.43, and the median absolute sharpness ratio is 0.047 ($\log_2$, about 3\%). SIFT matches within $\pm20^\circ$ of the equator (1,183 samples with at least ten matches) give a median absolute vertical offset of 0.99\,arcmin; 87.5\% of the samples stay below 3\,arcmin, one pixel at the 20\,px/deg analysis scale. The largest offsets occur where the matches themselves spread widely: 13 of the 14 samples with an offset of at least 10\,arcmin have a median absolute deviation of at least 5\,arcmin across their matches (1.3\,arcmin over all samples), suggesting parallax from near objects rather than misalignment; the exception has only 21 matches. Over samples, the 5th and 95th percentiles of horizontal disparity have medians of $0.14^\circ$ and $2.10^\circ$, reflecting the depth range of the scenes.

\begin{table}[t]
\centering\small
\caption{Left/right consistency over the 1,211 samples (one frame each, central $\pm45^\circ$ of half-EQ; absolute values). Vertical offset: 1,183 samples with at least ten equatorial matches.}
\label{tab:stereo}
\begin{tabular}{lccc}
\toprule
Measure & Median & 95th pct. & Max \\
\midrule
Luma difference (\%) & 0.40 & 2.49 & 8.57 \\
Mean-color $\Delta E$ & 0.43 & 1.28 & 2.54 \\
Sharpness ratio ($\log_2$) & 0.047 & 0.144 & 0.558 \\
Vertical offset (arcmin) & 0.99 & 4.11 & 25.1 \\
\bottomrule
\end{tabular}
\end{table}

\textbf{Geometry conversion.} Round trips on three scenes (both calibrations and eyes, 16-bit RGB, no lossy coding) are summarized in Table~\ref{tab:roundtrip}. The converter interpolates bilinearly by default; its Lanczos4 option preserves markedly more detail. Round trips are not lossless: 5.4--5.8\% of non-black native pixels fall outside the inverse-mapping domain, mostly in lateral edge bands. The converter's half-EQ also differs from Resolve's in viewing orientation (about 45\,px at $7200^2$ before alignment); a single spherical rotation fitted per calibration and eye, transferred to other scenes, leaves median feature residuals of 1.05--2.01\,px. The released converter does not apply this rotation. On an x86-64 Linux host, 9 functional tests and 16 native-resolution conversions (both calibrations, eyes and directions) passed.

\begin{table}[t]
\centering\small
\caption{Round-trip conversion quality over six eye-frames, on shared valid support excluding near-black pixels.}
\label{tab:roundtrip}
\begin{tabular}{llcc}
\toprule
Domain & Interp. & RGB PSNR (dB) $\uparrow$ & Luma SSIM $\uparrow$ \\
\midrule
Fisheye & Bilinear & 42.8--53.1 & 0.9906--0.9978 \\
Fisheye & Lanczos4 & 53.9--64.9 & 0.9994--0.9999 \\
Half-EQ & Bilinear & 45.3--54.8 & 0.9945--0.9983 \\
Half-EQ & Lanczos4 & 57.3--66.3 & 0.9997--0.9999 \\
\bottomrule
\end{tabular}
\end{table}

\begin{figure*}[t]
\centering
\includegraphics[width=\textwidth]{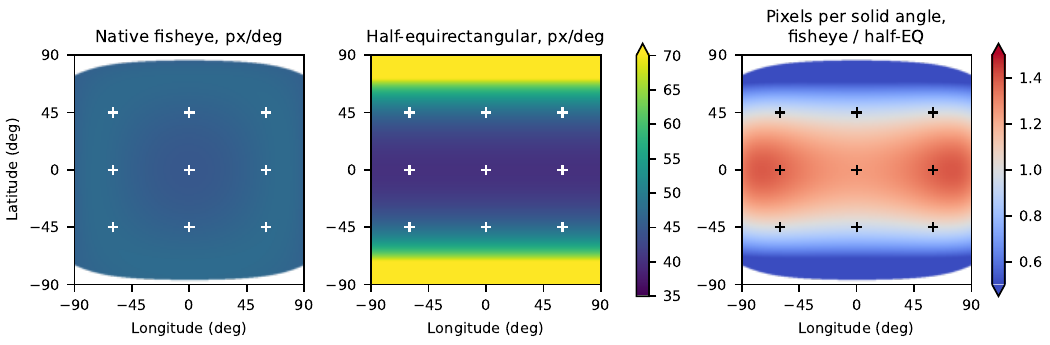}
\caption{Angular sampling density (pixels per degree) of both representations over the half-EQ domain, and the ratio of pixels per solid angle (fisheye / half-EQ). Crosses mark the centers of the nine evaluation viewports.}
\Description{Three maps over longitude and latitude from -90 to 90 degrees. The fisheye density is nearly uniform at about 45 pixels per degree; the half-equirectangular density is about 40 at the equator and increases towards the poles; the ratio map is above one in a band around the equator and below one near the top and bottom.}
\label{fig:density}
\end{figure*}

\section{Fisheye versus Half-EQ Coding}
\label{sec:study}
VR180 pipelines commonly encode the equirectangular layout, while the camera records fisheye. Because every sample is released in both forms, rendered from the same camera files, we ask: \emph{as delivered, which representation reaches a given viewport quality with less bitrate, and where in the field of view do they differ?}

\textbf{Clips.} Before any of them was encoded, we fixed 24 video clips by deterministic diversity selection over recording date, scene type, motion, brightness and calibration, taking at most one clip per overlap component. They cover 22 scene types, 19 recording dates and both cameras (14/10). From each clip and eye we use 90 frames (1\,s at 90\,fps, one GOP) from the middle of the clip.

\textbf{Sources.} The inputs are the released fisheye and half-EQ HEVC files, rendered in DaVinci Resolve at about 570 and 510\,Mb/s per eye, roughly ten times the highest test rate. They are decoded to 10-bit 4:2:0 and re-encoded without any conversion. Both renders are black outside the recorded field: the fisheye beyond about $180^\circ$ and the half-EQ near the zenith and nadir (Sec.~\ref{sec:construction}).

\textbf{Coding and evaluation.} Both representations are encoded with x265 4.1 (Main10, preset medium, 1\,s closed GOP, 8-thread pool) at CRF 18, 24, 30 and 36. Each decoded stream and its own source are rendered with Lanczos4 interpolation into nine $1536\times1536$, $60^\circ$ rectilinear viewports (yaw $\in\{-60^\circ,0^\circ,60^\circ\}$ $\times$ pitch $\in\{45^\circ,0^\circ,-45^\circ\}$), and Y$'$ PSNR (primary) and SSIM are computed on pixels that are valid and non-black in both sources. WS-PSNR~\cite{wspsnr} is not used because its weights are defined for equirectangular and cube-map layouts, not for a fisheye raster. The two eyes are coded independently, and the stereo rate is the sum of their video payloads. The rate difference at equal quality is the BD-rate~\cite{bdrate} with PCHIP interpolation over each clip's overlapping PSNR range (negative favors fisheye). Because each representation is compared with its own source, detail lost when Resolve projects the camera image to half-EQ is not counted; as a check, the mean luma gradient of the fisheye sources in the viewports is only 4\% higher than that of the half-EQ sources (median over clips; range 3--11\%).

\begin{figure}[t]
\centering
\includegraphics[width=\columnwidth]{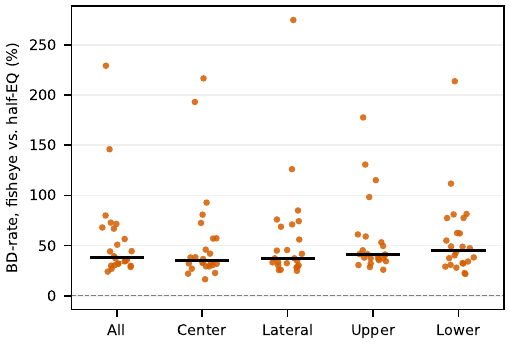}
\caption{Per-clip BD-rate of fisheye relative to half-EQ in each viewport region (black bar: median). Positive values mean fisheye needs more bitrate for the same viewport PSNR.}
\Description{A strip plot of per-clip BD-rate values in five viewport regions with median bars; all values are positive.}
\label{fig:bdrate}
\end{figure}

\textbf{Results.} Figure~\ref{fig:bdrate} and Table~\ref{tab:bdrate} report all 24 clips; every clip yields a BD-rate in every region. Over all nine viewports, the fisheye needs a median of $38.4\%$ more bitrate than half-EQ for the same viewport PSNR (range $+24.1\%$ to $+229.2\%$), and half-EQ needs less bitrate for every clip in every region, consistent with the finding of Eichenseer and Kaup for distortion-corrected fisheye coding~\cite{eichenseer2014}. The gap is smallest in the center viewport (median $+35.0\%$) and largest in the lower viewports ($+45.5\%$). At equal CRF the half-EQ streams reach 0.5--0.6\,dB higher median PSNR while the fisheye streams are 1.13--1.20 times larger. Per-clip rate--quality curves are in the code repository.

\begin{table}[t]
\centering\small
\caption{BD-rate of fisheye relative to half-EQ (\%) per region over the 24 clips. Available clips, median and range; clips favoring fisheye is exploratory.}
\label{tab:bdrate}
\begin{tabular}{lcccc}
\toprule
Region & Avail. & Median & Range & Fisheye better \\
\midrule
All nine & 24 & +38.4 & +24.1 to +229.2 & 0 \\
Center & 24 & +35.0 & +16.5 to +216.7 & 0 \\
Lateral & 24 & +37.4 & +24.8 to +274.8 & 0 \\
Upper & 24 & +40.9 & +26.0 to +177.7 & 0 \\
Lower & 24 & +45.5 & +21.8 to +213.8 & 0 \\
\bottomrule
\end{tabular}
\end{table}

\textbf{Sampling density.} To interpret these differences, we computed the angular sampling density of both representations in every viewing direction from the factory lookup tables (Fig.~\ref{fig:density}). The fisheye is sampled almost uniformly, from 44.6 pixels per degree on the optical axis to 47.1 at $75^\circ$ off axis. The half-EQ has 40 pixels per degree on the equator and becomes denser towards the poles. In the center viewport the fisheye therefore has 21\% more pixels per solid angle than the half-EQ (45.0 vs.\ 40.9 px/deg), in the upper and lower viewports about 9\% fewer (46.4 vs.\ 50.6), and over the nine viewports about the same. The BD-rate pattern does not follow pixel count: the fisheye needs more bitrate in every region, including the top and bottom where it has fewer pixels. Both representations sample about twice as densely as the evaluation viewports (23.2 px/deg at the viewport center), so the metric cannot reward detail above about 23 px/deg; a plausible explanation, not tested separately here, is that coding the strongly curved fisheye geometry is less efficient for block-based motion compensation and intra prediction than coding the straighter half-EQ geometry.

\textbf{Exploratory analyses.} The following analyses are descriptive. With SSIM in dB, the median BD-rate over all viewports is $+53.7\%$, with the same sign as with PSNR for all 24 clips. Grouped by camera motion, object motion, low light, low texture or camera, all group medians with at least two clips are positive ($+30.3\%$ to $+50.9\%$; $+50.5\%$ for the 12 low-light clips and $+35.4\%$ for the others). A bootstrap over the 24 clips gives a 95\% interval of $+32.1\%$ to $+56.6\%$ for the median. Because the 24 clips are a diversity selection rather than a random sample, these intervals are descriptive only.

\section{Ethics, Access and Limitations}
\label{sec:limitations}
\textbf{People and privacy.} The footage was recorded mostly in publicly accessible places. Bystanders, audiences and public performers may appear and some may be identifiable. We did not obtain individual consent from people incidentally captured, and we do not blur faces: the dataset's purpose is unmodified camera-native imagery, and BRAW frames cannot be anonymized without re-encoding. The data are released for non-commercial use (CC BY-NC 4.0) behind an access-request gate, the annotations contain no identity information, and identifying, tracking or profiling individuals is not an intended use. Anyone who appears in the data can request removal through the dataset repository; removed samples are excluded from later versions.

\textbf{Use of AI tools.} We disclose the use of AI in this research. Labels were generated with the DeepSeek V4 API and reviewed with GPT-6 Astra, and the event scene types were checked by Claude (Sec.~\ref{sec:annotation}); no label was verified against human ground truth. The AI coding assistants OpenAI Codex and Anthropic Claude Code were used to write the processing, validation and experiment code and to assist with data analysis. The clip selection was fixed before any clip was encoded; analyses beyond the regional BD-rate summary are labeled exploratory. All reported numbers come from the released scripts.

\textbf{Access.} Data: \url{https://huggingface.co/datasets/lulinxuan/VR180} (CC BY-NC 4.0, automatic access gate), with a datasheet~\cite{datasheets}. Code and lookup tables: \url{https://github.com/lulinxuan/vr180-dataset-tools}.

\textbf{Limitations.} All footage comes from two cameras of the same model and was manually selected, so coverage is biased: one aviation and defense exhibition contributes 28\% of the samples and only 18 clips have a moving camera. Labels are AI-generated from sparse frames. Factory calibration is released as provided and not independently measured, and it is not metric depth ground truth. Resolve render settings are only partly recoverable. Conversion between representations is lossy at the lateral edges and differs from Resolve in orientation. The coding study uses one encoder, 24 diversity-selected 1\,s clips and each representation's own render as reference, so detail lost in the projection to half-EQ is not counted; it codes the two eyes independently, without inter-view prediction such as MV-HEVC, compares only fisheye and half-EQ (no cube-map layouts), and its evaluation viewports sample more coarsely than either representation.

\section{Conclusion}
We release a camera-native stereo VR180 dataset that pairs BRAW with native-fisheye and half-equirectangular video, factory lens metadata, conversion tools and AI-generated annotations. Its paired representations and camera RAW support studies of representation choices that web-derived resources cannot. Re-encoding the released renders, native-fisheye coding needed a median 38\% more bitrate than half-equirectangular coding at equal viewport quality, for every clip and viewing direction. A second capture phase will add longer clips of night scenes, indoor daily life and close-range objects.

\bibliographystyle{ACM-Reference-Format}
\bibliography{refs}

\end{document}